\documentclass[runningheads]{llncs}
\usepackage{graphicx}
\usepackage{caption}
\usepackage{subcaption}
\usepackage{algorithmic}
\usepackage[ruled,vlined]{algorithm2e}
\usepackage{tabularx}
\usepackage{amsmath}
\usepackage{booktabs}
\usepackage[export]{adjustbox}
\usepackage{float}
\usepackage{comment}
\begin{document}
\title{
An Exploratory Study on Simulated Annealing for Feature Selection in Learning-to-Rank}
%
%
\author{Mohd. Sayemul Haque \and
Md. Fahim \and
\\Muhammad Ibrahim*
\orcidID{0000-0003-3284-8535}
}
\authorrunning{Saymul et al.}
%
\institute{Department of Computer Science and Engineering, University of Dhaka, Bangladesh 
\email{sh.sayem.haque36@gmail.com, fahimcse381@gmail.com, $^*$ibrahim313@du.ac.bd} (*Corresponding Author)\\
}

\maketitle              
\begin{abstract}
Learning-to-rank is an applied domain of supervised machine learning. As feature selection has been found to be effective for improving the accuracy of learning models in general, it is intriguing to investigate this process for learning-to-rank domain. In this study, we investigate the use of a popular meta-heuristic approach called simulated annealing for this task. Under the general framework of simulated annealing, we explore various neighborhood selection strategies and temperature cooling schemes. We further introduce a new hyper-parameter called the progress parameter that can effectively be used to traverse the search space. Our algorithms are evaluated on five publicly benchmark datasets of learning-to-rank. For a better validation, we also compare the simulated annealing-based feature selection algorithm with another effective meta-heuristic algorithm, namely local beam search. Extensive experimental results shows the efficacy of our proposed models. 

\end{abstract}
\section{Introduction}

Information Retrieval (IR)\cite{baeza1999modern}  deals with the methods, process and different procedures of searching, locating, and retrieving any kind of meaningful and structured information. These information may come from the database, documents or any kind of source like web. Indeed, it is also considered as a science of searching information where the search queries can be based on searching for meta data (data about data), searching for ranking the information, document lists or the information from the databases or documents. In an IR-based system, a query is given to the system and the job of the system is to retrieve the information from the documents or database which are most relevant to the query.

Generally an IR model is probabilistic or mathematical. IR usually deals with bigger datasets, such as a lot of documents like web pages. In this modern age of technology, the databases are becoming bigger and bigger and number of documents of any collection is increasing rapidly. Furthermore, variations of data are becoming available. The  probabilistic or mathematical IR models are not sufficient to handle these large and complex datasets as these models are slow and less accurate. So some automation techniques are needed in the IR system for doing the job quickly and accurately.

Machine Learning \cite{jordan2015machine} is the process of automated learning from data. These algorithms learn from the historical data and improve themselves without requiring any manipulation by the user.   As nowadays there is huge amount of data available, so it is easier to build a robust machine learning model. Learning-to-Rank (LtR) \cite{ibrahim2015_tf} is an application of the machine learning and information retrieval. It uses supervised machine learning to solve the ranking problem. The aim of LTR is to come up with optimal ordering of items given a query. The training data for an LTR model consist of a list of query-document pairs. The features of the training data are the output scores of various IR scoring functions of the documents (such as tf, idf, tf-idf, bm25 score etc.), and the labels are relevance scores which indicates how much relevant a document is with respect to a given query. These data are fed into a machine learning model. After the learning process is done based on the training data, the model is used to generate relevance scores for unseen query-document pairs \cite{ibrahim2022understanding}

Nowadays the size of data is increasing rapidly. Large amount of data is not only increasing the model complexity but also affecting the performance. In information retrieval, especially in web search, usually the data size is very large and thus training of ranking models is computationally costly. Besides, redundant and irrelevant features may lead to poor effectiveness of the models. So selecting a good subset of  feature is quite important for LtR field \cite{sajid2023feature}.

\subsection{Motivation}
For building a robust LtR model, we need to handle the redundant features and prevent the overfitting problem. A proper subset of features is needed not only for building a better LtR model but also for faster training. There are several methods for feature selection like filter method, wrapper method, embedded method and meta-heuristic approach.

Meta-heuristic algorithms are used to solve different large and complex optimization problems such as NP-hard problems. These algorithms provide better solutions to incomplete optimization problems and are easily adaptable to different circumstances. They also perform well in solving the feature selection problem of machine learning. Among the meta-heuristic approaches, simulated annealing is a choice of many researchers for its ability to find good solutions quickly. 

\subsection{Research Question}
The main objective of this investigation is to address the feature selection problem in learning-to-rank using an effective meta-heuristic algorithm, namely simulated annealing algorithm. In particular, the following research questions are addressed in this study: 

\begin{itemize}
    \item How do various neighbor generation strategies perform?
    \item Which cooling schemes are effective?
    \item How can the temperature be used to traverse the search space? 
\end{itemize}

\subsection{Contributions}
The following contributions are made in this research:
\begin{itemize}
    \item We adapt the standard simulated annealing algorithm to select a good subset of features in learning-to-rank domain.
    \item We utilize two effective neighbourhood definitions for simulated annealing.
    \item We explore the effects of different cooling schemes in simulated annealing algorithm. 
    \item We introduce a novel technique called the progress parameter in the basic simulated annealing algorithm to better traverse th search space.
    \item We compare the performance of all these settings of simulated annealing algorithm on five benchmark LtR datasets. To better validate our investigated techniques, we also compare these results with another meta-heuristic algorithm called local beam search algorithm.
\end{itemize}

The rest of the paper is organized as follows. In Section~\ref{sec:related_work}, we describe the related research works and identify the research gap. In Section~\ref{sec:proposed} we propose a framework for feature selection problem in learning-to-rank domain using simulated annealing algorithm. Here we discuss how we define the feature selection problem, the state definition, neighbours structure, cooling schemes, and progress parameter. In Section~\ref{sec:results} we discuss our experimental settings and analyze the results. Finally in Section~\ref{sec:conclusion} we conclude the paper mentioning some directions for future work.

\section{Background and Related Work}
\label{sec:related_work}
In this section, we briefly discuss the relevant existing research, thereby identifying the research gap in the existing literature. 

\subsection{Feature Selection in Supervised Machine Learning}
The quality of the features in a dataset has a major impact on the performance of machine learning models. While building a machine learning model for a real life scenario, it is rare that all the features in the dataset are useful to make prediction. Including redundant features reduces the generalization capability of the model. It may also lead to overfitting problem if the number of instances is comparatively low. Thus sometimes the redundant features may decrease the performance of the model. Feature selection techniques are used to choose a subset of the features from the original feature set by removing irrelevant, redundant and noisy features. 
Generally, careful use of feature selection methods leads to better performance of machine learning models \cite{guyon2003introduction}. 
These methods are briefly described  below.

\subsubsection{Traditional Methods}
The most widely used non-heuristic feature selection approaches are 
1. \textit{Filter Method},
2. \textit{Wrapper Method},
3. \textit{Embedded Method}.

\textit{Filter methods} pick up the intrinsic properties of the features measured using univariate statistics instead of cross-validation performance of learning models. As these methods do not use any learning model to evaluate performance and instead solely depends on statistical methods, they are computationally less expensive than other methods like wrapper and embedded methods. These methods are preferable when the feature space is too large \cite{ambusaidi2016building}. Several techniques like SelectKBest using Chi Square test, Fisher Score's, Correlation Coefficient, Variance threshold, ReliefF \cite{sanchez2007filter} etc. are examples of such methods.

\textit{Wrapper methods} employ a machine learning model to evaluate the potential subsets of features \cite{kohavi1995feature}. These potential subsets are generated using a search method. After selecting a subset, it is used to train the model that generates predictions for validation dataset. Based on this prediction, the subset is evaluated using performance metrics. The search methods help exploring the feature space efficiently. This is an iterative and computationally expensive process but it is usually more accurate than the filter methods \cite{maldonado2009wrapper}. There are few techniques used in wrapper method like forward feature selection, backward feature elimination, exhaustive feature selection, recursive feature elimination etc \cite{kohavi1995feature}.

\textit{Embedded methods} make a trade-off between computational cost and accuracy by combining the wrapper and filter methods, which often results in better accuracy while keeping the computational cost reasonable \cite{chandrashekar2014survey}. Embedded methods employ an iterative process in the sense that they take care of each iteration of the model's training process and carefully extracts those features which contribute the most to the training error for a particular iteration. Regularization methods are the most commonly used embedded methods which penalize a feature given a coefficient threshold. These methods are far less computationally intensive than wrapper methods \cite{liu2005toward}.

\subsubsection{Meta-Heuristic Approaches}
Meta-heuristics are problem-independent techniques \cite{sharma2020comprehensive}, i.e., they do not include problem specific features. They are very efficient to avoid the local optima problem in optimization problems since, unlike greedy or purely heuristic approaches, they may accept a worse solution while maneuvering in the search space. Meta-heuristic optimization algorithms are widely used to solve large-scale optimization problems \cite{rajpurohit2017glossary}. The advantages of the these algorithms over traditional optimization algorithms are their adaptability and the ability of dealing with complex problems \cite{yang2009harmony}. 

Like many domains, meta-heuristic algorithms work well in the feature selection problem of machine learning. Although these approaches are slow, they are often able to select near-optimal feature subsets \cite{lin2008parameter}. So many researchers have used different meta-heuristic approaches for solving the feature selection problem. Jihoon Yang and Vasant Honavar et al. \cite{yang1998feature} use one of the popular meta-heuristic algorithm called genetic algorithm in this regard. After finding best ``parents'' from the ``population'', new generations (a.k.a ``child'') are created by crossing over the parents. Then a mutation process is conducted to bring some new property in the child. Lucija Brezoˇcnik et al. \cite{brezovcnik2018swarm} use Swarm Intelligence \cite{dario2012robots} techniques for solving feature selection problem. Here after initializing population, the agents are modified or updated based on a fitness function until it reaches a termination criterion. Another popular meta-heuristic algorithm is simulated annealing. Siedlecki et al. \cite{siedlecki1993automatic} use simulated annealing algorithm for automatic feature selection. Simulated annealing algorithm tries to find optimal solution by exploring almost the whole search space. This algorithm is used in some other feature selection technique like Hybrid Whale Optimization \cite{mafarja2017hybrid}. Alvi et al. \cite{allvi2020feature} use simulated annealing algorithm for feature selection, but in a very naive setting and with a very few datasets. 

\subsection{Feature Selection in Learning-to-Rank}
The number of features determines the time required to extract features from documents, both in the development phase and testing phase. Also, inclusion of noisy features reduces the accuracy of model. Some research have been performed on feature selection for LtR task. In \cite{geng2007feature}, the authors propose a filter-based method, namely Greedy Search Algorithm (GAS) where the problem is treated as an optimization problem. The authors try to find features with maximum total importance scores and minimum total similarity scores. In \cite{gigli2016fast}, the authors propose three filter-based feature selection methods, namely NGAS, XGAS and HCAS. In NGAS, the authors first add the most relevant feature to the set. After that goes on a loop adding features which minimize similarity and maximizes relevance. XGAS is an upgraded version of NGAS that compares the feature to be added to a subset of features. HCAS performs a hierarchical agglomerative clustering. But this type of feature selection does not ensure the optimal feature selection as argued in \cite{das2001filters}. In \cite{sousa2019risk},the authors propose a wrapper method which is a single multi objective criteria for feature selection. In \cite{pan2011greedy}, authors propose an evolutionary algorithm-based strategy that reduces the search space by eliminating weak features during the traversing process. Although this type of strategy is adaptive and ensures near-optimal solutions, wrapper methods are time consuming. FSMRank \cite{lai2013fsmrank} is an embedded strategy which simultaneously reduces the ranking error along with feature selection. In \cite{rahangdale2019deep}, a deep neural network-based architecture is used to automatically select features that uses l1 regularization. 

\subsection{Research Gap}
As we have discussed above, a lot of research has been done to investigate the impact of feature selection on learning-to-rank domain which mostly consists of filter methods. However, apart from the genetic algorithm, there has not been any significant work on the impact of different meta-heuristic algorithms. In the current research, we intend to investigate the efficacy of a popular meta-heuristic algorithm, namely simulated annealing, to select a useful and informative set of features in the LtR domain.

\subsection{Simulated Annealing Technique}
Since our work is based on simulated annealing, in this sub-section we briefly describe this procedure.

Simulated annealing \cite{van1987simulated} is one of the most popular and widely used optimization algorithms. This algorithm is used to solve different real life and interesting problems  like travelling salesman problem, scheduling problems, task allocations, graph colouring and partitioning, non-linear function optimizations and so on \cite{vidal1993applied}. It uses a probabilistic technique for approximating the global optimum of a given function. Simulated annealing gets its motivation from the process of slow cooling of metals. It performs better than bare greedy algorithms because of its ability to overcome the local optimum problem. This process is very useful for situations where there are a lot of local minima \cite{bertsimas1993simulated}.  

\subsubsection{Simulated Annealing Algorithm}
Simulated annealing algorithm takes the idea of the physical annealing process where a hot metal is shaped by gradually reducing the temperature. The algorithm uses this temperature cooling scheme for optimizing parameters in a model \cite{kirkpatrick1983optimization}. Simulated annealing algorithm performs well because it does not only explore the better states but also traverses some worse states (aka bad moves). If we only look for the better states (like traditional hill climbing method), we may get stuck in a local optima. In contrast, if we allow to traverse some apparently bad moves, we are able to explore a larger search space which helps in finding the global optima. However, if we allow an unrestrained amount of bad moves, the solution may get worsened. Thus simulated annealing algorithm trades off between this two extremes. Before a bad move is about to be taken place, an acceptance probability is calculated. If the acceptance probability is less than some random threshold value, the next state (worse state) is not explore. Otherwise, the bad move is explored. The acceptance probability is usually calculated as $e^{- \Delta E/ T}$ where $T$ is the temperature and $\Delta E$ the  difference of solution qualities between the two states in question.  We see that when the higher the temperature, the higher the acceptance probability, so the algorithm is more likely to accept a bad state. This means that the algorithm is more likely to explore bad moves when the temperature is higher. Thus the temperature has a significant effect on performance of simulated annealing algorithm  \cite{kirkpatrick1983optimization}. The temperature is controlled by annealing or cooling schedule which helps to reduce the value of temperature gradually from a higher value to the lower one.

The pseudo code (\cite{russell2002artificial}) of generic simulated annealing algorithm is given in Algorithm~\ref{algo:SA}. Note that this generic algorithm must be adapted differently to different problem domain.

\begin{algorithm}
\SetAlgoLined
\KwResult{Returns a solution state }
\textbf{Input}: $problem$, a problem\;
\textbf{Input}: $schedule$, a mapping from time to temperature\;
$current \gets$ MAKE-NODE ($problem.InitialState$)\;
\For{t = 1 to $\infty$ }{
$T$ $\gets$ schedule($t$)\;
\If{T = 0} {return $current$\;}
next  $ \gets$ a randomly selected successor, i.e., neighbor of current\;
$\Delta E \gets next.Value - current.Value$\;
\eIf{$\Delta E > 0$}
{$current \gets next$}
    {$prob$ = $e^{\Delta E/ T}$\;
     \If{$prob > rand(0,1)$}{
       $current \gets next$
    }
   }
}
 \caption{Generic Simulated Annealing Algorithm \cite{russell2002artificial}}
\label{algo:SA}
\end{algorithm}

\section{Proposed Framework}
\label{sec:proposed}

In this section we elaborately discuss our proposed framework of simulated annealing-based feature selection for learning-to-rank which includes discussions on the definition of solution state, evaluation of a solution, neighborhood definition, traversing strategy, and annealing or cooling schemes. 
Below in Algorithm~\ref{algo:SA FS existing} we provide a generic framework as to how the simulated annealing algorithm can be used to solve feature selection problem in learning-to-rank \cite{kuhn2019feature}. This generic version will later be modified followed by our discussions on specific details of the algorithm.

\begin{algorithm}
\SetAlgoLined
\KwResult{Returns best subset of features of size $k$. ($k$ is provided as a hyper-parameter.)}
 create an initial random subset of features of size $k$, $current$\;
 \For{t $\gets$ 1 to $\infty$}{ 
 T $\gets$ schedule(t)\;

 \If{T = 0} {return $current$\;}
 a new feature subset, $next$ $\gets$ $get\_neighbour(current)$\;
 fit model with $next$ features and estimate performance of learnt model\;
  $\Delta E \gets $ performance($next$) - performance($current$)\;
  \eIf{$\Delta E > 0$}{
   accept $next$ as the new current subset, i.e., $current \gets next$\;
   }{
   $probability \gets e^{\Delta E/ T}$\;
   \eIf{probability $>$ rand(0,1)}
   {accept $next$ as the new current subset, i.e., $current \gets next$\;}
   {reject the new subset, i.e., no change in $current$\;}
  }
 }
 \caption{Simulated Annealing-Based Framework for Feature Selection in Learning-to-Rank}
 \label{algo:SA FS existing}
\end{algorithm}

\subsection{Adapting Generic Simulated Annealing Algorithm to Feature Selection in Learning-to-Rank}
In order to use simulated annealing for feature selection in learning-to-rank scenario, we need to decide on several aspects of the algorithm which are as follows:
\begin{itemize}
    \item How to define a state?
    \item How to evaluate the quality of a state?
    \item How to define the neighborhood relationship among the states?
    \item How to traverse the search space?
    \item How to decide on the annealing or cooling scheme?
\end{itemize}

Below we elaborately discuss these aspects which leads to our proposed framework.

\subsubsection{Definition of a State}
In the feature selection problem of learning-to-rank, we intend to search for the subset of features that gives us the best performance in terms of IR metrics like NDCG or MAP \cite{ibrahim2015_tf}. So a state in our setting represents a subset of features among the available ones.
To implement this idea, we take a bit array whose length is equal to number of total available features. Each bit represents a feature, and $1$ in $i$-th index means that we include $i$-th feature for measuring IR performance, and $0$ means otherwise. 
We search for the best subset among the states with $k$ number of features, where $k$ ranges from 1 to $n-1$, where $n$ is the total number of features. For each subset of features for each $k$, we invoke simulated annealing and assign annealing steps in proportion to the the total number of states and the number of neighbors of a state. The reason behind this approach is to get a better view of the impact of increasing number of features.

\subsubsection{Quality of a State}
To measure the quality of a state in simulated annealing, we need to consider the task at hand, which is learning-to-rank. In this domain, the quality of the list of documents/elements  ranked by an algorithm is measured against some IR metrics like NDCG, MAP etc. Therefore, for each state, i.e., a subset of features among the available ones, we train the model learnt with only the features that are included in the feature subset at hand. We then measure the performance of the learnt model using NDCG and MAP.

\subsubsection{Definition of Neighbourhood}
One of the most important things for building a robust simulated annealing algorithm is to define the neighbouring states. We borrow two effective neighbourhood definitions from existing literature. These two neighbour selection techniques have been used in the greedy hybrid operator proposed by Zhen et al. \cite{zhan2016list} and Wang et al. \cite{wang2015solving} to produce candidate solutions from the current state. We use a slightly modified version of these techniques. The two neighbours selection techniques are:  0's and 1's Swapping and Insertion.

i. \textbf{\emph{0's and 1's Swapping}}: 
Here a state's neighbors are defined as those having one feature different from the current state. Specifically, considering two sets, where one holds the indexes of the bit-array (current state) with 1's and the later with 0's. From each set we randomly choose an index and flip the bits (swap) of those indexes to generate neighbors. Suppose, our current state is $\phi$. Two indexes $i$ and $j$ are chosen randomly from the sets of 1's and 0's respectively. After that, by performing $flip(\phi[i])$ and $flip(\phi[j])$ we generate a neighbour state.

ii. \textbf{\emph{Insertion}}: Here we choose two random positions $i$ and $j$ from the current state. We then  move the element in position $j$ to position $i$ and shift all values from $i$ by 1 position to the right. In case of $i < j$, if our current state is $\phi$ then we generate a neighbour state $\phi\prime$ where 
$\phi\prime[i+1]$ =  $\phi[i]$ for all $i$ up to $j$, and $\phi\prime[i]$ = $\phi[j]$. The other case, i.e., $i > j$, works similarly.

\begin{figure}[h]
\begin{subfigure}{0.40\textwidth}
\includegraphics[width=0.9\linewidth, height=5cm]{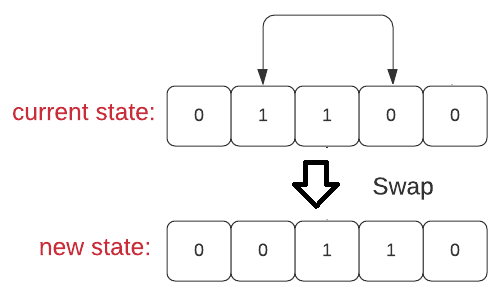} 
\caption{Swap}
\label{fig:swap}
\end{subfigure}
\begin{subfigure}{0.6\textwidth}
\includegraphics[width=0.9\linewidth, height=5cm]{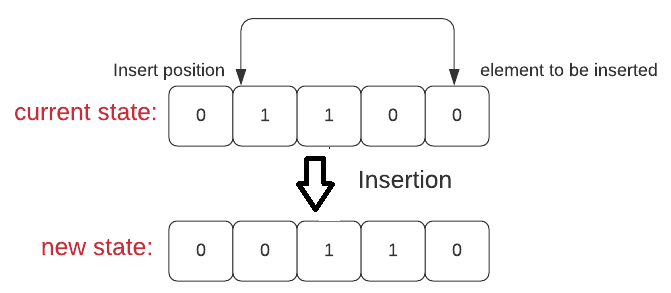}
\caption{Insertion}
\label{fig:mq8_n2}
\end{subfigure}
\caption{Two neighborhood definitions.}
\label{fig:neighbor}
\end{figure}

Fig.~\ref{fig:neighbor} pictorially demonstrates the two neighboring schemes. 

\subsubsection{Traversing the Neighborhood: Progress Parameter}
The problem with selecting a neighbor randomly from the set of neighbors is that we may miss potentially good states. So Connolly \cite{connolly1990improved} proposes to evaluate neighbors in a sequential order. However, in our problem the number of neighbors is large. So instead of visiting every neighbor sequentially, we introduce a novel parameter which we call the ``progress'' parameter. This parameter keeps track of the number of iterations of the algorithm that has not seen update on the current best solution. When the progress parameter reaches a threshold value, we restart from the current best solution. This way it is likely to help in avoiding the local optima. It also ensures exploring the current best solution's neighborhood space thoroughly in lower temperatures so that it, at least, ensures the local optima of that neighborhood space if not the global optima.

\subsubsection{Annealing Scheme}
An important aspect of simulated annealing algorithm is the temperature \cite{vidal1993applied}. We not only traverse the better states but also some worse states to find a good global solution. 
Traditionally, the probability score based on which a bad move is accepted or not is measured by function called \textit{metropolice}, which is calculated as $e^{\Delta E/ T}$,  where $\Delta E = next.Value - current.Value$ \cite{russell2002artificial}. We see that traversing 
a worst state directly depends on the temperature -- if the temperature is high, so is the probability score, and hence there is a better chance for it to be accepted, and vice versa. 

The algorithm starts with a bigger value of temperature ($T_{initial}$) and then slightly cools down as per a cooling schedule. As a result, in the early iterations of the algorithm, we allow more bad moves, and this allowance is gradually restricted as the algorithm matures. Thus the performance of simulated directly depends on the cooling schedule. There are several cooling schemes (aka annealing schemes) in the existing literature. Among them, we employ three schemes which are popular and widely used. These schemes are described below:

i. \textbf{\emph{Geometric Annealing Scheme}}: 
A popular yet simplest cooling scheme is geometric annealing scheme \cite{cohn1999simulated}. This scheme uses the following rule to update the temperature: $T_{t+1} = \alpha * T_{t}$, where  $t$ is the time step, $T_{t}$ and $T_{t+1}$ are the current and updated temperatures, respectively, and $\alpha$ is a controlling parameter. Generally,  $\alpha$ is in (0,1]. For our experiments, we set $\alpha$ = 0.9.

ii. \textbf{\emph{Logarithmic Annealing Scheme}}:
Another effective cooling scheme is logarithmic annealing scheme. Though this scheme is slow, it is reported to perform better than some other schemes in the smaller problems \cite{cohn1999simulated}. Here the temperature update rule is: $T_{t} = \frac{T_0}{\log \left(t+t_{0}\right)}$, where $t$ is the time step, $T_{0}$ is the initial temperature, $T_{t}$ is the updated temperature at $t$-th time step and $t_{0}$ is a  fixed value to prevent division by zero. For our experiments, we use $t_{0}$ = 10.

iii. \textbf{\emph{Fast Annealing Scheme}} : 
Harold Szu et al. \cite{szu1987fast} introduce a Fast Simulated Algorithm (FSA) which is claimed to be much faster than the classical simulated annealing. In this paper a new temperature update rule is introduced, which is: $T_{t} = \frac{T_{0}}{(1+t)}$, where $t$ is the time step, $T_{0}$ is the initial temperature, $T_{t}$ is the updated temperature at $t$-th time step.

\subsubsection{Temperature Length}
The temperature length parameter is the number of steps after which the temperature is updated using the cooling scheme. States are evaluated at certain temperatures until it is reduced to a terminating value by the cooling schedule. There are some options such as fixed temperature length, adaptive method and variable length method. We choose the adaptive method proposed by Abramson \cite{abramson1991constructing} where the temperature is updated depending on the search progress. When the number of accepted moves crosses a certain value, the temperature is updated, otherwise, the same temperature is used in the next iteration.

\subsection{Implementation of Simulated Annealing Algorithm for Feature Selection in Learning-to-Rank}

\vspace{8pt}
\begin{algorithm}
\SetAlgoLined
\KwResult{Returns best subset of features of size $k$. ($k$ is provided as a hyper-parameter.)}
 \textit{current-state} $\gets$ select-initial-state()//a subset of features of size $k$\;
 $T \gets T_{initial}$\;
  \textit{best-state} $\gets$ \textit{current-state}\;
 \For{t $\gets$ 1 up to no. of iterations}{
 \If{T = 0} {return \textit{current-state}\;}
 \textit{next-state} $\gets$ 
 $get\_neighbour($\textit{current-state}$)$ \;
  Fit LtR model on training data using features expressed by \textit{next-state}\;
  //Evaluating performance on test data\;
  $\Delta E \gets $ Evaluation(\textit{next-state}) $-$ Evaluation(\textit{current-state})\;
  \eIf{$\Delta E > 0$}{
    \textit{current-state} $\gets$ \textit{next-state}\;
   \If{$ Evaluation($\textit{next-state}$) > Evaluation($\textit{best-state}$)$}{
   Update $\textit{best-state}$ using $\textit{next-state}$}
   }{
   $probability$ = $metropolice(\Delta E$)\;
   \eIf{probability $>$ rand(0,1)}
   {
       \textit{current-state} $\gets$ \textit{next-state}\;
   }
   {Reject \textit{next-state}, i.e., no change in \textit{current-state}\;}
  }
  \If{Temperature update condition is met}{
    $T \gets$ cooling-scheme($t$) \;
  }
  \If{Progressing condition is not met}{
  Restart from the \textit{best-state}
  }
 }
 \caption{Proposed Simulated Annealing Algorithm for Feature Selection in Learning-to-Rank}
 \label{algo:SA proposed}
\end{algorithm}
Based on all the ideas and concepts discussed above, we devise the algorithm  for solving the feature selection problem in learning-to-rank which is given in Algorithm~\ref{algo:SA proposed}. Here our contributions over the traditional simulated annealing are in four places: neighborhood generation, cooling scheme, temperature length, and restart traversing based on local history of state updates (using progress parameter).

\subsection{Local Beam Search}

Although in this research our main objective is to investigate efficacy of various settings of simulated annealing algorithm for selecting a useful feature subset in LtR task, for the sake of better validation of our proposed framework, we adapt another meta-heuristic algorithm called local beam search to the same problem, and compare the experimental results. The authors in \cite{dash1997feature}, \cite{aha1996comparative} show the use of LBS algorithm for feature selection in machine learning.

Local Beam Search (LBS) is a popular meta-heuristic algorithm \cite{freitag2017beam} which is well-known for its efficient memory usage \cite{ow1988filtered}. Whereas the heuristic algorithms (like hill climbing) consider only one successor of the states at a time, LBS keeps track on more than one successors at a time. At every level of search tree, it considers $\beta$ number of successors where $\beta$ is called the beam-width. For large search spaces where there is insufficient amount of memory to store the fully grown search tree, the LBS algorithm thus offers help by keeping track a small number of successors. The algorithm uses breadth-first search while traversing  a search tree. While building the tree, the algorithm generates all the successors of the states at each level. It then sorts the successors in an increasing order of their heuristic values. It then stores only the top $\beta$ successors. The memory requirement is bounded by the value of $\beta$ to perform the search. At the end of the search, the algorithm returns the best successor as a solution from the last $\beta$ successors. Algorithm~\ref{algo:LBS proposed} shows the pseudocode of the algorithm we have devised for solving feature selection problem in learning-to-rank  using local beam search algorithm.

\vspace{8pt}
\begin{algorithm}
\SetAlgoLined
\KwResult{Returns best subset of features of size $k$. ($k$ is provided as a hyper-parameter.)}
 $q$ $\gets$ beam-width\;
 \textit{current-top-q-best-states} $\gets$ select-$q$-initial-states()//subsets of size $k$\;
 \For{t = 1 to q}{
 \textit{state} $\gets$ \textit{current-top-q-best-states}[t] \;
 Fit LtR model on training data with the feature set represented by $state$\;
 //Evaluate performance on test data\;
 \textit{performance-scores[t]} $\gets$ \textit{Evaluation}(\textit{state})\;
 }
 \textit{best-state} $\gets$  $ -\infty$ \;
 
 \For{t = 1 to no. of steps}{
 \textit{current-top-q-best-states} $\gets$ sort-descending-order(performance-scores)\;
  \For{t = 1 to q}{
   \textit{current-state} $\gets$ \textit{current-top-q-best-states}[t]\;
   \textit{new-state} $\gets$ $Neighbours($\textit{current-state}$)$\;
   Fit LtR model on training data with the feature set represented by \textit{new-state}\;
   //Evaluate performance on test data\;
   \If{ \textit{Evaluation}$($\textit{new-state} $)$ $>$ \textit{Evaluation}$($\textit{current-state}$)$}{
   $Update($\textit{current-top-q-best-states, performance-scores}$)$}
   }
 }
\textit{current-top-q-best-states} $\gets$ sort-descending-order(performance-scores)\;
\textit{best-state} $\gets$ \textit{current-top-q-best-states}[1]\;
\caption{Proposed Local Beam Search-Based Algorithm for Feature Selection in Learning-to-Rank}
\label{algo:LBS proposed}
\end{algorithm}

\section{Experiments and Result Analysis}
\label{sec:results}
In this section we discuss the datasets and experimental setup, and analyze the results.

\subsection{Datasets}
For our experiments we use five benchmark LETOR datasets as described below:
\begin{itemize}
    \item \textbf{MQ2008:} MQ2008 \cite{qin2013introducing} is part of Microsoft LETOR 4.0 package, which was released in July, 2009. There are 784 queries. Each query-document pair is labeled using a relevance score (0, 1 or 2). Each row consists of 46 features, along with query ID, comment about the document-query pair, document ID and the relevance score. 
    \item \textbf{MQ2007:} MQ2007 \cite{qin2013introducing} is also part of the LETOR 4.0 package. The dataset structure is similar to the MQ2008 (46 features). But there are 1692 queries in this dataset. 
    \item \textbf{OHSUMED:} Oregon Health Sciences University’s MEDLINE Data Collection also known as OHSUMED \cite{hersh1994ohsumed}. It is a subset of the MEDLINE dataset. It consists of 45 features and a total of 106 queries. It also contains multiple labels (0, 1 or 2).
    \item \textbf{TD2004:} Topic distillation 2004 aka. TD2004 \cite{qin2010letor} is one of the seven datasets of the LETOR 3.0 collection. It has 44 features and a total of 107 queries.
    \item \textbf{MSLR-10K:} Microsoft released MSLR-WEB10K, a large dataset. The dataset has 136 features and 10000 queries. It gives relevance score to each query-document pair from a scale of one to five (0, 1, 2, 3 or 4). 
\end{itemize}

\subsection{Rank-Learning Model}
In order to evaluate the ranking performance of the model trained on the selected feature subsets, we employ a widely-used and state-of-the-art rank-learning algorithm called LambdaMART~\cite{burges2010ranknet}.

\subsection{Evaluation Metrics}
We evaluate the ranking efficacy of a model using two metrics: Normalized Discounted Cumulative Gain (NDCG) and Mean Average Precision (MAP) \cite{ibrahim2015_tf}. NDCG is the ratio of Discounted Cumulative Gain (DCG) score of the ranking of items predicted by the model to that of ideal ranking. MAP is the mean of all the average precision (AP) scores for each query. 

\subsection{Algorithm Setup}
For simulated annealing, in Table~\ref{tab:different settings notations}, the two neighboring strategies are noted by \textbf{n1} and \textbf{n2} and the three cooling schemes are noted by \textbf{s1},\textbf{s2} and \textbf{s3}. So in total, there are six settings. For local beam search algorithm, we set \textit{beam-width} to be 10.

\begin{table}[h]
\caption{Various Settings of Simulated Annealing}
\begin{tabular}{|m{0.1\textwidth}|m{0.8\textwidth}|}
\hline
\multicolumn{1}{|c|}{\textbf{Abbreviation}} & \multicolumn{1}{c|}{\textbf{Description}}                  \\ \hline
n1s1    & neighborhood selection strategy: swap (n1), cooling scheme: geometric (s1)  with a cool-down factor of 0.9      \\ \hline
n1s2    & neighborhood selection strategy: swap (n1), cooling scheme: logarithmic (s2)                                  \\ \hline
n1s3    & neighborhood selection strategy: swap (n1), cooling scheme: fast annealing (s3)                 \\ \hline
n2s1    & neighborhood selection strategy: insertion (n2), cooling scheme: geometric (s1)  with a cool-down factor of 0.9\\ \hline
n2s2    & neighborhood selection strategy: insertion (n2), cooling scheme: logarithmic (s2)                          \\ \hline
n2s3    & neighborhood selection strategy: insertion (n2), cooling scheme: fast annealing (s3)                          \\ \hline
\end{tabular}
\label{tab:different settings notations}
\end{table}

\subsection{Experimental Results}
 In what follows, we demonstrate the following experimental results for each dataset:
\begin{itemize}
    \item We first compare different cooling strategies for each neighboring technique for simulated annealing.
    \item We then compare the ranking performance with progress parameter with ranking performance without progress parameter for simulated annealing.
    \item We then show the ranking performance of local beam search and compare it with that of simulated annealing. 
\end{itemize}
In each case, we report NDCG@10 and MAP scores. As our algorithms are non-deterministic, we take the average of these scores over ten iterations. Since extensive experiments on the (large) MSLR-WEB10K is computationally highly expensive, we discuss its experiments separately at a later stage of our analysis.

\subsubsection{Combinations of Neighborhood Strategies and Cooling Schemes}

\begin{figure}[H]
\begin{subfigure}{0.5\textwidth}
\includegraphics[width=0.9\linewidth, height=5cm]{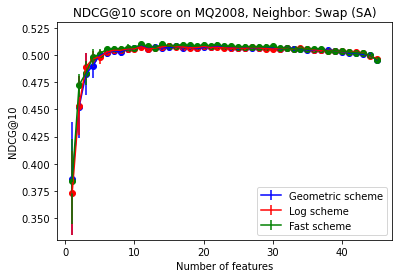} 
\caption{Neighborhood strategy: Swap}
\label{fig:mq2008_sa_ndcg_n1}
\end{subfigure}
\begin{subfigure}{0.5\textwidth}
\includegraphics[width=0.9\linewidth, height=5cm]{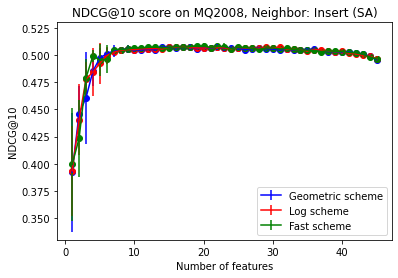}
\caption{Neighborhood strategy: Insertion}
\label{fig:mq2008_sa_ndcg_n2}
\end{subfigure}
\begin{subfigure}{0.5\textwidth}
\includegraphics[width=0.9\linewidth, height=5cm]{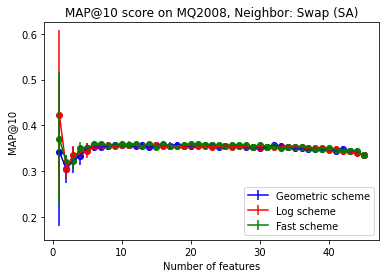} 
\caption{Neighborhood strategy: Swap}
\label{fig:mq2008_sa_map_n1}
\end{subfigure}
\begin{subfigure}{0.5\textwidth}
\includegraphics[width=0.9\linewidth, height=5cm]{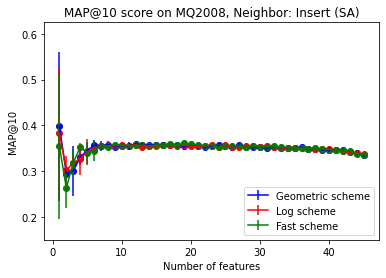}
\caption{Neighborhood strategy: Insertion}
\label{fig:mq2008_sa_map_n2}
\end{subfigure}
\caption{NDCG@10 and MAP scores on MQ2008 dataset for Simulated Annealing}
\label{fig:mq2008_sa_ndcg}
\end{figure}

\textbf{\emph{MQ2008 Data}}: Fig. \ref{fig:mq2008_sa_ndcg} plots the feature selection performance of simulated annealing evaluated using NDCG and MAP metrics for all the six settings. In these graphs, $x$-axis represents the number of features and the $y$-axis represents the corresponding NDCG@10 scores. 
For swapping, fast annealing reaches the highest average NDCG score of $0.50970$ among the cooling strategies. For insertion also, fast annealing reaches a peak of $0.5082$. For both graphs the standard error is small enough for the results to be accepted after number of selected features is more than 8.  
The maximum MAP score $0.3599$ is achieved by fast annealing for swapping. For insertion, a maximum MAP score of $0.3605$ is achieved by fast annealing.

\begin{figure}[H]
\begin{subfigure}{0.5\textwidth}
\includegraphics[width=0.9\linewidth, height=5cm]{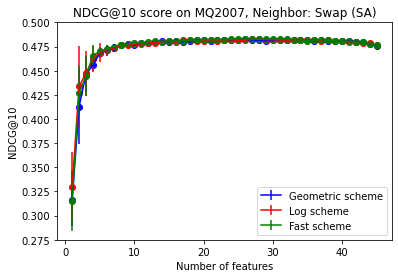} 
\caption{Neighborhood strategy: Swap}
\label{fig:mq2007_sa_ndcg_n1}
\end{subfigure}
\begin{subfigure}{0.5\textwidth}
\includegraphics[width=0.9\linewidth, height=5cm]{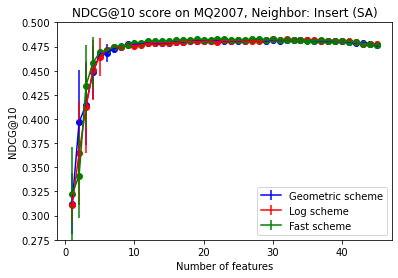}
\caption{Neighborhood strategy: Insertion}
\label{fig:mq2007_sa_ndcg_n2}
\end{subfigure}
\begin{subfigure}{0.5\textwidth}
\includegraphics[width=0.9\linewidth, height=5cm]{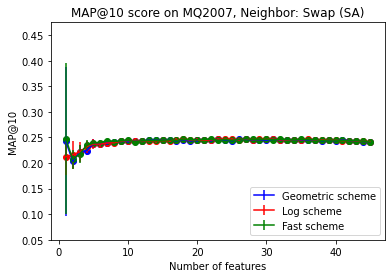} 
\caption{Neighborhood strategy: Swap}
\label{fig:mq2007_sa_map_n1}
\end{subfigure}
\begin{subfigure}{0.5\textwidth}
\includegraphics[width=0.9\linewidth, height=5cm]{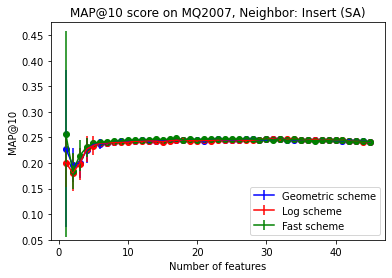}
\caption{Neighborhood strategy: Insertion}
\label{fig:mq2007_sa_map_n2}
\end{subfigure}
\caption{NDCG@10 and MAP scores on MQ2007 dataset for Simulated Annealing}
\label{fig:mq2007_sa_ndcg}
\end{figure}
\textbf{\emph{MQ2007 Data}}:
Fig. \ref{fig:mq2007_sa_ndcg} plots the feature selection performance of simulated annealing evaluated using NDCG score on MQ2007 for all the six settings. 
For swapping, fast annealing reaches the highest average NDCG score of $0.4829$ among the cooling strategies. For insertion, fast annealing also reaches a peak of $0.4828$. Both neighbor selection strategies give almost similar result in this case. Swapping works slightly better compared to insertion. For both graphs the standard error is small enough for the result to be accepted after number of selected features is more than 6. 
The maximum MAP score of $0.2478$ is achieved by fast annealing for swapping. For insertion, a maximum map score of $0.2486$ is also achieved by fast annealing.

\begin{figure}[H]
\begin{subfigure}{0.5\textwidth}
\includegraphics[width=0.9\linewidth, height=5cm]{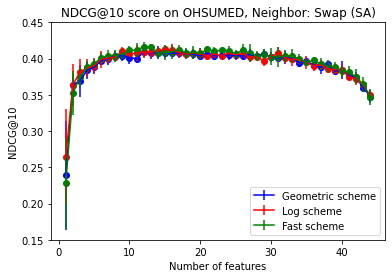} 
\caption{Neighborhood strategy: Swap}
\label{fig:OHSUMED_sa_ndcg_n1}
\end{subfigure}
\begin{subfigure}{0.5\textwidth}
\includegraphics[width=0.9\linewidth, height=5cm]{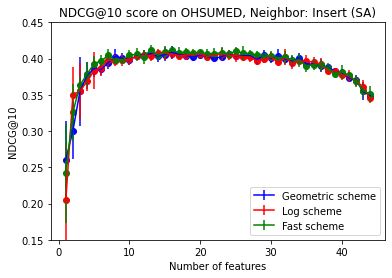}
\caption{Neighborhood strategy: Insertion}
\label{fig:OHSUMED_sa_ndcg_n2}
\end{subfigure}
\begin{subfigure}{0.5\textwidth}
\includegraphics[width=0.9\linewidth, height=5cm]{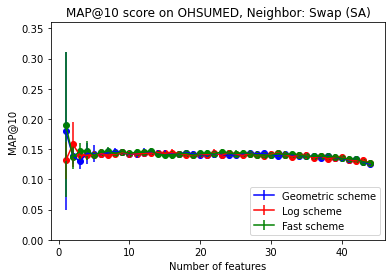} 
\caption{Neighborhood strategy: Swap}
\label{fig:OHSUMED_sa_map_n1}
\end{subfigure}
\begin{subfigure}{0.5\textwidth}
\includegraphics[width=0.9\linewidth, height=5cm]{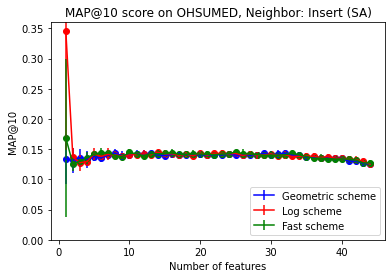}
\caption{Neighborhood strategy: Insertion}
\label{fig:OHSUMED_sa_map_n2}
\end{subfigure}
\caption{NDCG@10 and MAP scores on OHSUMED dataset for Simulated Annealing}
\label{fig:OHSUMED_sa_ndcg}
\end{figure}

\textbf{\emph{OHSUMED Data}}: Fig. \ref{fig:OHSUMED_sa_ndcg} plots the feature selection performance of simulated annealing evaluated using NDCG score on OHSUMED dataset for all the six settings. 
For swapping, fast annealing reaches the highest average NDCG score of $0.4165$ among the cooling strategies. For insertion, fast annealing also reaches a peak of $0.4121$. Swapping is found to be more effective compared to insertion. For both graphs the standard error is small enough for the result to be accepted after number of selected features is more than 8. 
The maximum MAP score of $0.1455$ is achieved by fast annealing for swapping. For insertion, a maximum map score of $0.1448$ is achieved by logarithmic scheme.

\begin{figure}[H]
\begin{subfigure}{0.5\textwidth}
\includegraphics[width=0.9\linewidth, height=5cm]{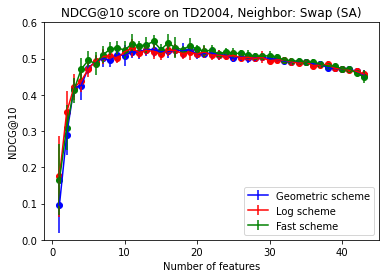} 
\caption{Neighborhood strategy: Swap}
\label{fig:TD2004_sa_ndcg_n1}
\end{subfigure}
\begin{subfigure}{0.5\textwidth}
\includegraphics[width=0.9\linewidth, height=5cm]{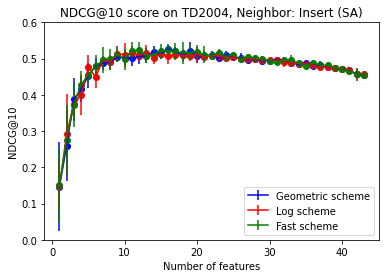}
\caption{Neighborhood strategy: Insertion}
\label{fig:TD2004_sa_ndcg_n2}
\end{subfigure}
\begin{subfigure}{0.5\textwidth}
\includegraphics[width=0.9\linewidth, height=5cm]{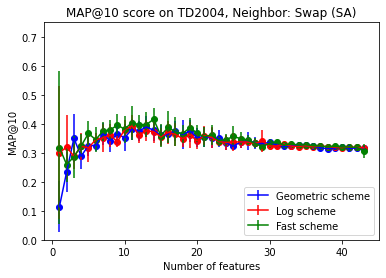} 
\caption{Neighborhood strategy: Swap}
\label{fig:TD2004_sa_map_n1}
\end{subfigure}
\begin{subfigure}{0.5\textwidth}
\includegraphics[width=0.9\linewidth, height=5cm]{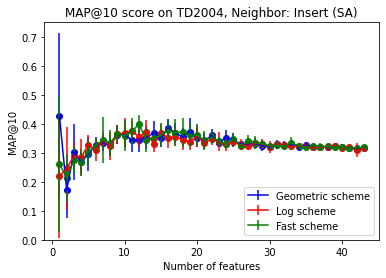}
\caption{Neighborhood strategy: Insertion}
\label{fig:TD2004_sa_map_n2}
\end{subfigure}
\caption{NDCG@10 and MAP scores on TD2004 dataset for Simulated Annealing}
\label{fig:TD2004_sa_ndcg}
\end{figure}
\textbf{\emph{TD2004 Data}} : Fig. \ref{fig:TD2004_sa_ndcg} plots the feature selection performance of simulated annealing evaluated using NDCG score on TD2004 dataset for all the six settings.  
For swapping, fast annealing reaches the highest average NDCG score of $0.5474$ among the cooling strategies. For insertion, fast annealing also reaches a peak of $0.5227$. Swapping is more effective compared to insertion. For both graphs the standard error is small enough for the result to be accepted after number of selected features is more than 6. 
The maximum MAP score of $0.4160$ is achieved by fast annealing for swapping. For insertion, a maximum MAP score of $0.3977$ is achieved by fast annealing.

\subsubsection{Effect of Progress Parameter}

\begin{figure}[H]
\begin{subfigure}{0.5\textwidth}
\includegraphics[width=0.9\linewidth, height=5cm]{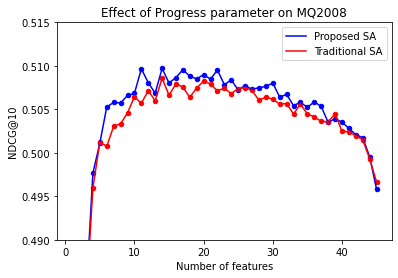} 
\caption{Progress parameter: MQ2008}
\label{fig:prog MQ2008}
\end{subfigure}
\begin{subfigure}{0.5\textwidth}
\includegraphics[width=0.9\linewidth, height=5cm]{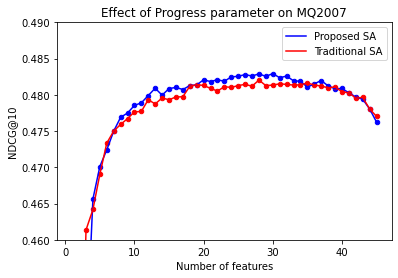}
\caption{Progress parameter: MQ2007}
\label{fig:prog MQ2007}
\end{subfigure}
\begin{subfigure}{0.5\textwidth}
\includegraphics[width=0.9\linewidth, height=5cm]{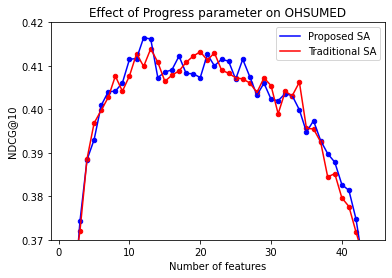} 
\caption{Progress parameter: OHSUMED}
\label{fig:prog ohsumed}
\end{subfigure}
\begin{subfigure}{0.5\textwidth}
\includegraphics[width=0.9\linewidth, height=5cm]{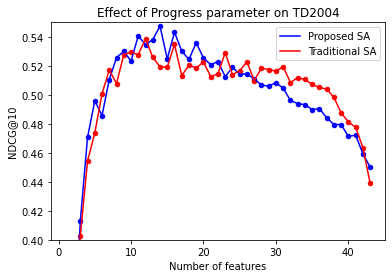}
\caption{Progress parameter: TD2004}
\label{fig:prog TD2004}
\end{subfigure}
\caption{Effect of progress parameter on performance.}
\label{fig:progress_parameter_alldata}
\end{figure}

Fig.~\ref{fig:progress_parameter_alldata} shows the comparison of our algorithm with progress parameter with the traditional one where the progress parameter is not included. As the swapping and fast annealing combination yielded the best results in our previous experiments, here the comparison is made using this setting. We can see that in most of the cases the traditional algorithm (i.e., without progress parameter) is outperformed by our proposed technique. 




\subsubsection{Local Beam Search Compared to Simulated Annealing }
Now we discuss the feature selection performance of local beam search, which is followed by the comaprison between local beam search and simulated annealning. 

\begin{figure}[H]
\begin{subfigure}{0.5\textwidth}
\includegraphics[width=0.9\linewidth, height=5cm]{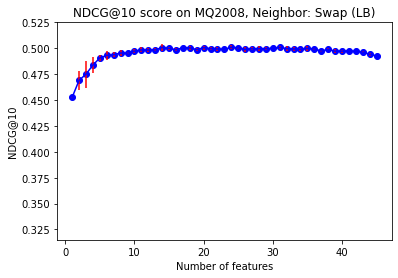} 
\label{fig:mq2008_lb_ndcg}
\caption{NDCG@10}
\end{subfigure}
\begin{subfigure}{0.5\textwidth}
\includegraphics[width=0.9\linewidth, height=5cm]{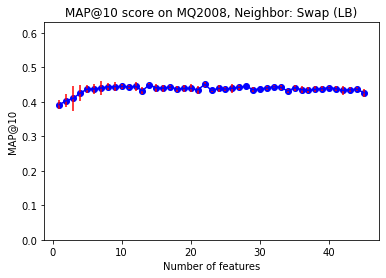}
\label{fig:mq2008_lb_map}
\caption{MAP@10}
\end{subfigure}
\caption{Ranking scores on MQ2008 dataset for local beam search}
\label{fig:mq2008_lb}

\begin{subfigure}{0.5\textwidth}
\includegraphics[width=0.9\linewidth, height=5cm]{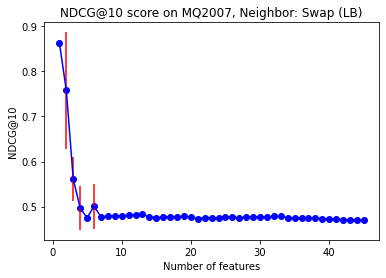} 
\label{fig:mq2007_lb_ndcg}
\caption{NDCG@10}
\end{subfigure}
\begin{subfigure}{0.5\textwidth}
\includegraphics[width=0.9\linewidth, height=5cm]{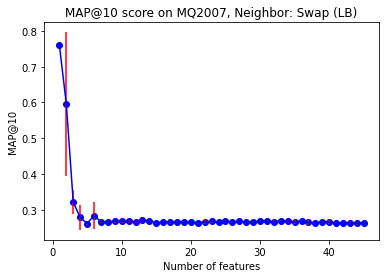}
\label{fig:mq2007_lb_map}
\caption{MAP@10}
\end{subfigure}
\caption{Ranking scores on MQ2007 dataset for local beam search}
\label{fig:mq2007_lb}
\end{figure}

\begin{figure}
\begin{subfigure}{0.5\textwidth}
\includegraphics[width=0.9\linewidth, height=5cm]{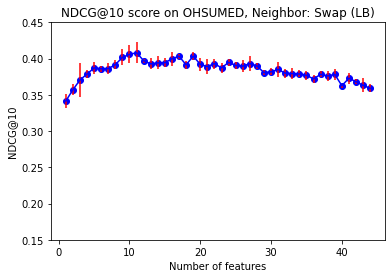} 
\label{fig:OHSUMED_lb_ndcg}
\caption{NDCG@10}
\end{subfigure}
\begin{subfigure}{0.5\textwidth}
\includegraphics[width=0.9\linewidth, height=5cm]{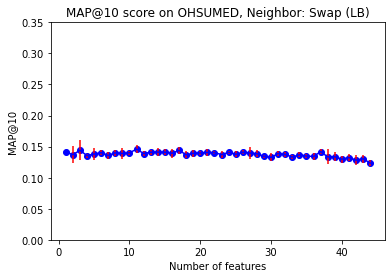}
\label{fig:OHSUMED_lb_map}
\caption{MAP@10}
\end{subfigure}
\caption{Ranking scores on OHSUMED dataset for local beam search}
\label{fig:OHSUMED_lb}

\begin{subfigure}{0.5\textwidth}
\includegraphics[width=0.9\linewidth, height=5cm]{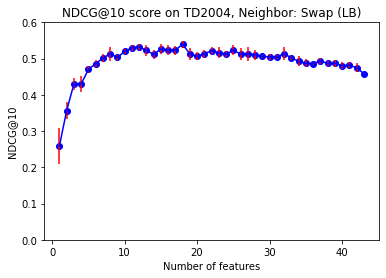} 
\label{fig:TD2004_lb_ndcg}
\caption{NDCG@10}
\end{subfigure}
\begin{subfigure}{0.5\textwidth}
\includegraphics[width=0.9\linewidth, height=5cm]{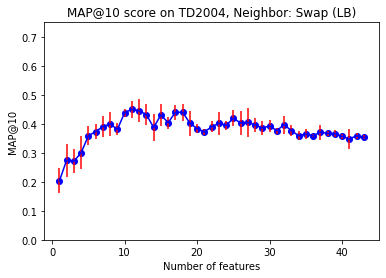}
\label{fig:TD2004_lb_map}
\caption{MAP@10}
\end{subfigure}
\caption{Ranking scores on TD2004 dataset for local beam search}
\label{fig:TD2004_lb}

\end{figure}

For MQ2008 dataset (Fig. \ref{fig:mq2008_lb}), the max NDCG score reached by local beam search is $0.5009$ and MAP score of $0.4503$. For MQ2007 dataset (Fig. \ref{fig:mq2007_lb}), the maximum NDCG and MAP scores are $0.4827$ and  $0.2703$ respectively. For OHSUMED dataset (Fig. \ref{fig:OHSUMED_lb}), the maximum NDCG score is $0.4081$ and MAP score is $0.1459$. For TD2004 dataset (Fig. \ref{fig:TD2004_lb}), the max NDCG score is $0.5398$ and MAP score of $0.4399$.

\begin{figure}
\begin{subfigure}{0.5\textwidth}
\includegraphics[width=0.9\linewidth, height=5cm]{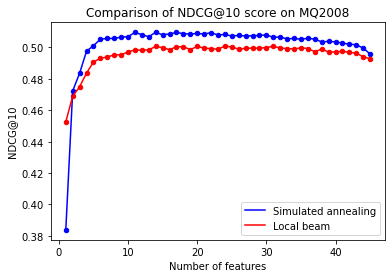} 
\label{fig:MQ2008_vs}
\caption{MQ2008}
\end{subfigure}
\begin{subfigure}{0.5\textwidth}
\includegraphics[width=0.9\linewidth, height=5cm]{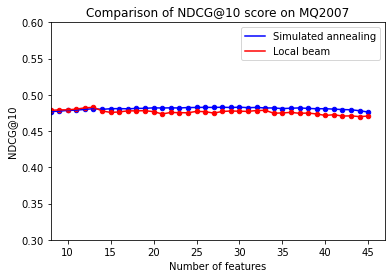}
\label{fig:MQ2007_vs}
\caption{MQ2007}
\end{subfigure}

\begin{subfigure}{0.5\textwidth}
\includegraphics[width=0.9\linewidth, height=5cm]{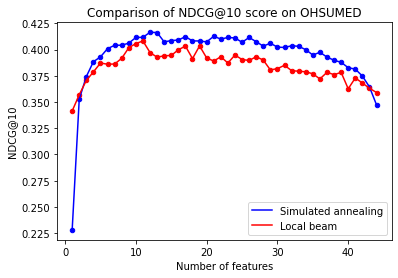} 
\label{fig:OHSUMED_vs}
\caption{OHSUMED}
\end{subfigure}
\begin{subfigure}{0.5\textwidth}
\includegraphics[width=0.9\linewidth, height=5cm]{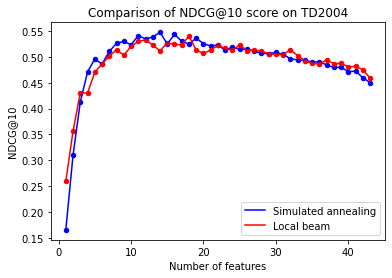}
\label{fig:TD2004_vs}
\caption{TD2004}
\end{subfigure}
\caption{Simulated annealing vs local beam search on all datasets}
\label{fig:all_vs}
\end{figure}

Fig. \ref{fig:all_vs} compares  simulated annealing (neighbor selection: swap, cooling scheme: fast annealing) and local beam search for all  nfour datasets. For MQ2008 dataset, although at very early stages, performance is better for local beam search, simulated annealing performs better as we increase the number of features. For MQ2007 dataset, both performs quite similarly, with simulated annealing having slightly better results. For OHSUMED and TD2004 datasets, simulated annealing performs much better in finding better feature subsets than local beam search.

\subsubsection{MSLR-WEB10K Data}
\begin{figure}[H]
\begin{subfigure}{0.5\textwidth}
\includegraphics[width=0.9\linewidth, height=5cm]{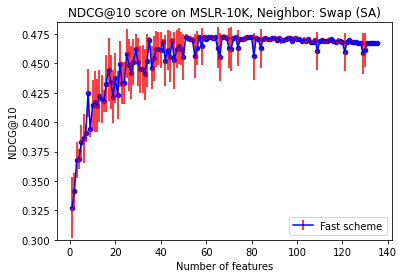} 
\label{fig:mslr10k_sa_ndcg}
\caption{NDCG@10}
\end{subfigure}
\begin{subfigure}{0.5\textwidth}
\includegraphics[width=0.9\linewidth, height=5cm]{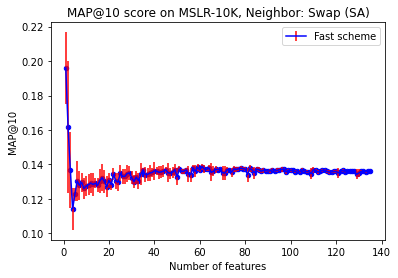}
\label{fig:mslr10k_sa_map}
\caption{MAP@10}
\end{subfigure}
\caption{Ranking scores on MSLR-WEB10K dataset for simulated annealing}
\label{fig:mslr10k_sa}
\end{figure}

\begin{figure}[H]
\begin{subfigure}{0.5\textwidth}
\includegraphics[width=0.9\linewidth, height=5cm]{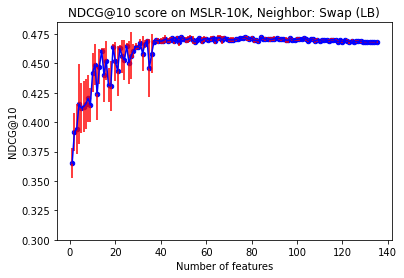} 
\label{fig:mslr10k_lb_ndcg}
\caption{NDCG@10}
\end{subfigure}
\begin{subfigure}{0.5\textwidth}
\includegraphics[width=0.9\linewidth, height=5cm]{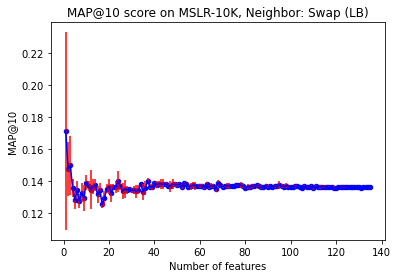}
\label{fig:mslr10k_lb_map}
\caption{MAP@10}
\end{subfigure}
\caption{Ranking scores on MSLR-WEB10K dataset for local beam search}
\label{fig:mslr10k_lb}
\end{figure}

\begin{figure}[h]
\includegraphics[height=6cm,center]{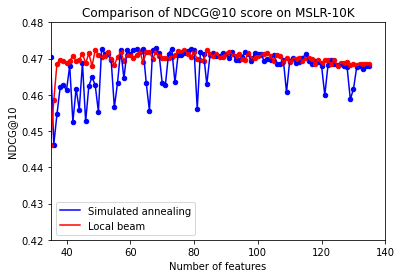}
\caption{Simulated annealing vs local beam search on MSLR-WEB10K dataset}
\label{fig:mslr10k_vs}
\end{figure}

The MSLR-WEB10K dataset is quite large compared to other datasets we have experimented with. So for this dataset we consider the best variant (neighbor selection: swap, cooling scheme: fast annealing) of simulated annealing found on the other four datasets. Also, for this dataset we conduct only one run of the algorithms instead of 10 runs. Fig. \ref{fig:mslr10k_sa} plots the feature selection performance of simulated annealing. The maximum NDCG score achieved by simulated annealing is $0.4729$ and MAP score of $0.1387$. Fig. \ref{fig:mslr10k_lb} plots the feature selection performance of local beam search. The maximum NDCG score reached by local beam search is $0.4725$ and MAP score of $0.1400$. Fig. \ref{fig:mslr10k_vs} compares between simulated annealing and local beam search. For the MSLR-WEB10K dataset, simulated annealing and local beam search both perform quite similarly.

\par

\subsection{Training Time Comparison}
\begin{figure}[H]
\includegraphics[height=7cm,center]{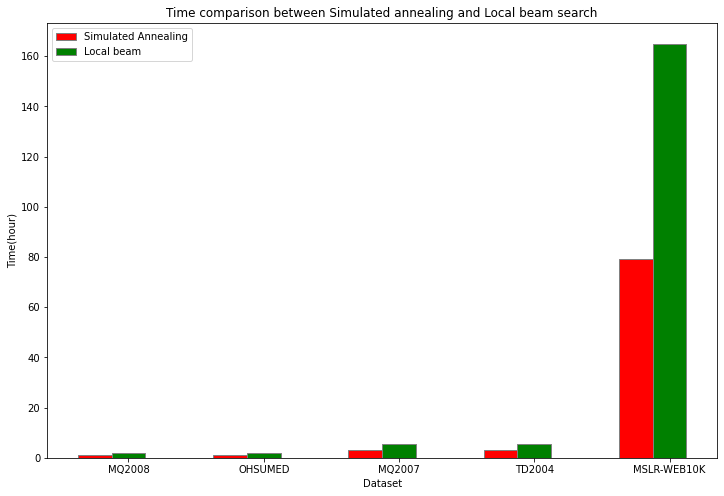}
\caption{Time comparison between simulated annealing  and local beam search}
\label{fig:time}
\end{figure}

Fig. \ref{fig:time} shows the training time comparison between simulated annealing and local beam search. Here the $x$ and $y$ axes represent the datasets and the required time (in hours) respectively. We see that local beam search is found to be computationally more expensive than simulated annealing for all five datasets.

\subsection{Discussion}
After analyzing the experimental results of the benchmark datasets, we observe the following patterns:
\begin{itemize}
    \item Feature selection improves the overall ranking accuracy of a model, and also reduces training time to some extent. This is because adding too many features may result in overfitting (especially if the number of training instances is relatively less) and increased training time. In all the datasets of our experiments, when the number of features was too low, the ranking performance was found to be unacceptable due to suffering from underfitting problem. As we kept on adding features, the ranking error and standard error started to go down up to a certain point. After that, adding more features did not improve the performance much. Instead, in some cases it resulted in lower performance, which, we conjecture, was due to overfitting problem or due to irrelevant or noisy features. The plots of NDCG@10 of OHSUMED and TD2004 datasets showed that adding too many features may even deteriorate the performance.
    \item In most scenarios we observe that the swap neighboring scheme seems to perform better than the insertion technique for simulated annealing. This is probably due to the fact that the insertion neighboring scheme is quite random in nature as inserting from one position to another shifts the whole set of selected features. Due to its randomness, it might give better result than swapping when cooling time is small compared to the problem. Whereas, swapping only exchanges one feature for another. As it may swap a feature that is contributing less to the performance of the model for a feature contributing more, this definition of neighbourhood enables us to find better solutions in a relatively controlled manner given sufficient cooling time, as it takes small steps towards a better solution.
    \item Among the cooling strategies, fast annealing is found to work better. Fast annealing converges faster than  the other two cooling schemes. As a result, when in the early iterations of simulated annealing, fast annealing scheme tends to give better result. However, in the very early iterations, fast annealing works similarly as the other two strategies.
    \item The progress parameter is found to improve the ranking accuracy across all datasets. This is because it helps the algorithm escape local optima as well as ensuring a better solution state among the neighbors of the current best state.
    \item Local beam search is found to be less effective than simulated annealing. We conjecture that the reason behind this is, local beam search always considers better solutions than the current one, thereby being more prone to getting stuck at a local optima. Also, its time requirement is larger as it keep a list of states of the beam length, and in each iteration, it needs to find the best state from the list. 
\end{itemize}

\section{Conclusion}
\label{sec:conclusion}
In applications where large number of instances are not available, feature selection may  significantly improve ranking performance of learning-to-rank algorithms. In this research, we have adapted an efficient meta-heuristic algorithm called simulated annealing to select a better subset of features for learning-to-rank problem.  We have investigated two neighbour selection approaches, three temperature and the cooling schemes, and also have incorporated a novel parameter for better traversal of the search space. We also compared the performance of our simulated annealing algorithms with another meta-heuristic algorithm called local beam search.

Several research directions have emerged from this research. Other meta-heuristic approaches may be investigated for solving the feature selection in learning-to-rank. It is necessary to  conduct further experiments on larger datasets. Further experimentation may be conducted to find the best settings of hyper-parameters to yield better ranking performance. Finding ways to reduce training time by carefully narrowing down the search space can also be an important investigation. 

%
%
%
%
\bibliographystyle{plain}
\bibliography{paper}
\end{document}